# Bias-constrained multimodal intelligence for equitable and reliable clinical AI

Cheng Li[1,7], Weijian Huang[1,6,7], Jiarun Liu[1,2,3], Hao Yang[1,2,3], Qi Yang[4], Song Wu[5], Ye Li[1], Hairong Zheng[1], Shanshan Wang[1]*

[1]Paul C. Lauterbur Research Center for Biomedical Imaging, Shenzhen Institutes of Advanced Technology, Chinese Academy of Sciences, Shenzhen, China.

[2]Pengcheng Laboratory, Shenzhen, China.

[3]University of Chinese Academy of Sciences, Beijing, China.

[4]Department of Radiology, Beijing Chaoyang Hospital, Capital Medical University, Beijing, China.

[5]Department of Urology, South China Hospital, Medical School, Shenzhen University, Shenzhen, China.

[6]Huawei Technologies Co., Ltd., Shanghai, China.

[7]These authors contributed equally: Cheng Li, Weijian Huang.

*e-mail: ss.wang@siat.ac.cn.

The integration of medical imaging and clinical text[1,2] has enabled the emergence of generalist artificial intelligence (AI) systems for healthcare[3,4]. However, pervasive biases, such as imbalanced disease prevalence, skewed anatomical region distributions, heterogeneous imaging protocols, and demographic disparities, pose significant challenges to the fairness and reliability of vision-language systems in real-world clinical settings[5,6]. Here we present BiasCareVL, a bias-aware multimodal learning framework that introduces bias control directly into model design, rather than treating it as a post hoc correction. BiasCareVL incorporates adaptive uncertainty modeling with optional human-in-the-loop refinement to regulate the influence of dominant data patterns and to promote equitable reasoning under distributional imbalance. Trained on 3.44 million samples spanning over 15 imaging modalities, the framework supports diverse clinical tasks, including visual question answering, disease classification, segmentation, and report generation within a unified representation space. Across eight public benchmarks covering dermatology, oncology, radiology, and pathology, BiasCareVL consistently outperforms 20 state-of-the-art methods, with pronounced gains in clinically challenging scenarios, including over 10% accuracy improvement in multi-class skin lesion diagnosis and more than 20% Dice improvement in small tumor segmentation. Furthermore, BiasCareVL achieves diagnostic performance exceeding human accuracy with substantially reduced time requirements when evaluated with board-certified radiologists. By open-

**sourcing BiasCareVL, we aim to promote a transparent, reproducible, and equitable future for AI in healthcare, paving the way for general-purpose, trustworthy, and clinically reliable AI systems.**

Recent breakthroughs in medical artificial intelligence (AI) have demonstrated that deep learning models can not only match but occasionally surpass human expertise in specific domains[7–9]. This rapid progress has paved the way for a new paradigm: the emergence of generalist medical AI[4,10]. In contrast to earlier approaches which were confined to single modalities or narrow tasks, these next-generation models are increasingly designed to integrate heterogeneous data streams, including radiological images, clinical notes, and laboratory results. Such integration enables a more comprehensive representation of patient health, more closely reflecting the complexity of real-world clinical decision-making[1,3,4,11,12].

A pivotal advancement in this trajectory is the development of medical vision-language pre-training[1–3,13,14]. By aligning visual representations with clinical semantics within a shared embedding space, vision-language models (VLMs) have unlocked the potential for unified, multi-task learning. These models can seamlessly perform visual question answering, generate radiology reports, and classify diseases across diverse anatomical regions, all within a single framework[3]. This convergence of imaging and text represents a significant leap toward scalable, general-purpose diagnostic systems that are better aligned with the open-ended and dynamic nature of clinical environments.

The shift toward these generalist paradigms signifies a maturation of AI in medicine, moving beyond task-specific applications to support comprehensive decision-making[1,4,11,12].

Despite these remarkable strides, a critical challenge persists on the path to truly reliable clinical AI: the pervasive issue of data bias[5,6]. Clinical datasets are inherently imbalanced, often reflecting historical inequities and logistical constraints rather than true population distributions[15,16]. These biases manifest at multiple levels: from modality availability bias, in which certain imaging techniques (e.g., MRI and CT) are overrepresented compared to others (e.g., PET)[17], to disease prevalence bias, where common conditions dominate the data while rare but high-risk pathologies remain scarcely represented[9], to demographic bias, reflected in persistent disparities in data distributions across age and racial groups[18]. Such imbalances skew model optimization toward frequent patterns, leading to inflated performance on aggregate benchmarks but potentially catastrophic failures in diagnosing underrepresented or rare conditions[5,19]. Consequently, addressing these latent biases is not merely a technical necessity but an ethical imperative for ensuring patient safety[16].

While recent studies have begun to explore bias mitigation, many existing approaches remain confined to vision-only pipelines or narrowly scoped tasks[20–23]. There is a distinct lack of unified solutions that effectively address bias within the context of multi-modal, generalist frameworks. This gap highlights the need for strategies that go beyond simple data reweighting or augmentation, demanding instead a robust integration of bias-aware mechanisms into the core architecture of vision-language systems[11].

Here we develop BiasCareVL, a fully open bias-constrained medical vision-language framework designed to enhance fairness and generalization in multimodal clinical AI (Fig. 1). BiasCareVL incorporates adaptive uncertainty modeling and human-in-the-loop refinement to mitigate skewed influence of dominant data patterns, ensuring robust performance in complex and ambiguous clinical situations. The framework is trained on a large-scale dataset of 3.44 million samples, consisting of 1.19 million image-text pairs and 2.25 million image samples, across more than 15 imaging modalities. This dataset exhibits substantial real-world biases, reflecting inherent imbalances in clinical data, including modality availability bias (Fig. 1a), anatomical prevalence bias (Fig. 1b), disease prevalence bias (Fig. 1c), and demographic bias (Fig. 1d-f).

We performed extensive evaluation on a wide range of downstream tasks, including visual question answering (VQA), multi-class and multi-label disease classification, organ and lesion segmentation, and report generation (Fig. 1g). BiasCareVL achieves superior performance compared to 20 state-of-the-art methods. In a multi-label chest disease classification task, BiasCareVL exceeds human diagnostic accuracy while substantially reducing the time required, as evaluated against three board-certified radiologists. Additionally, when assisted by BiasCareVL, human expert performance demonstrates clearly improvement. By releasing all code, data pipelines, and pretrained weights, we aim to promote transparency, reproducibility, and equitable progress in medical AI. Ultimately, we expect that BiasCareVL can lay a robust foundation for developing trustworthy, generalizable, and human-centric vision-language systems for real-world clinical deployment.

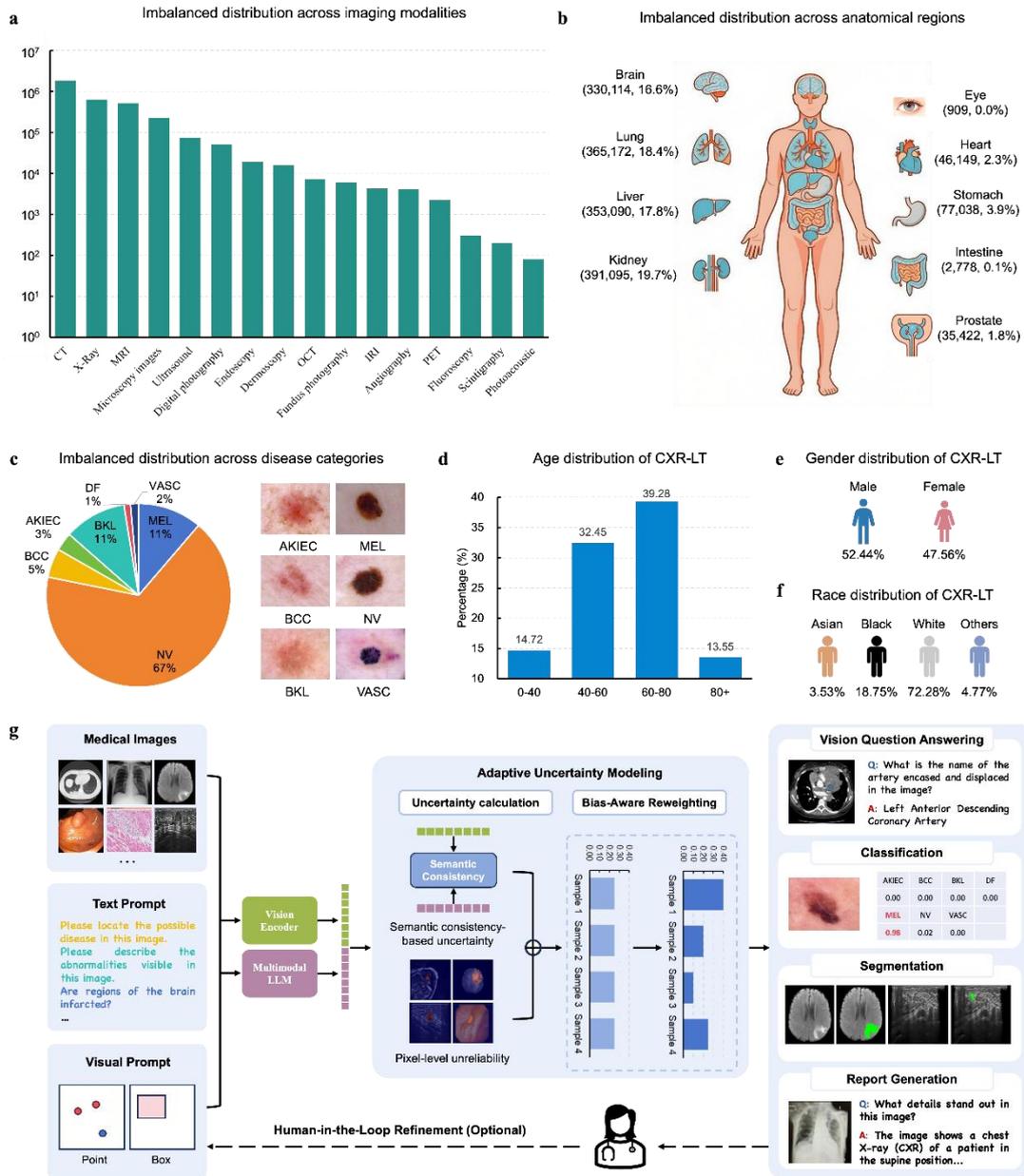

**Fig. 1 | BiasCareVL can promote equitable reasoning under distributional imbalance. a-f,** Illustrative examples of data biases prevalent in medical imaging,

including modality availability bias across the entire training dataset, anatomic prevalence biases in image segmentation tasks, disease prevalence biases in skin disease classification, and demographic information biases (age, gender, and race) in chest disease classification. **g**, The framework of BiasCareVL, depicting its adaptive uncertainty modeling and optional human-in-the-loop refinement for bias mitigation, and the spectrum of tasks it supports. CT: Computed tomography. MRI: Magnetic resonance imaging. PET: Positron emission tomography. OCT: Optical coherence tomography. IRI: Infrared reflectance imaging. AEIEC: Actinic keratosis/Bowen's disease (intraepithelial carcinoma). BCC: Basal cell carcinoma. BKL: Benign keratosis (solar lentigo/seborrheic keratosis/lichen planus-like keratosis). DF: Dermatofibroma. MEL: Melanoma. NV: Melanocytic nevus. VASC: Vascular lesion. LLM: Large language model.

**Closed-ended visual question answering**

To evaluate the diagnostic reasoning capability of BiasCareVL's MLLM pathway, we conducted closed-ended VQA experiments on two established benchmarks: VQA-RAD[24] and PathVQA[25]. As shown in Fig. 2a, general-purpose VLMs, including Qwen-VL-Chat[26], Qwen-VL2.5[27], DeepSeek-VL2[28], and RadFM[29], demonstrate limited performance on these medical benchmarks. Recent medical VLMs, such as BioMedGPT[3] and MedPLIB[30], show substantially stronger results, confirming the value of biomedical-aligned pretraining. Our proposed BiasCareVL achieves the best performance across all metrics on both datasets, outperforming both general-purpose and medical VLM baselines. Specifically, BiasCareVL improves accuracy by 1.2% on VQA-RAD and 1.8% on PathVQA over the second-best method, BioMedGPT. Compared to MedPLIB, which also reports multiple metric results, BiasCareVL achieves gains exceeding 10% on VQA-RAD and 1.2% on PathVQA across the reported scores.

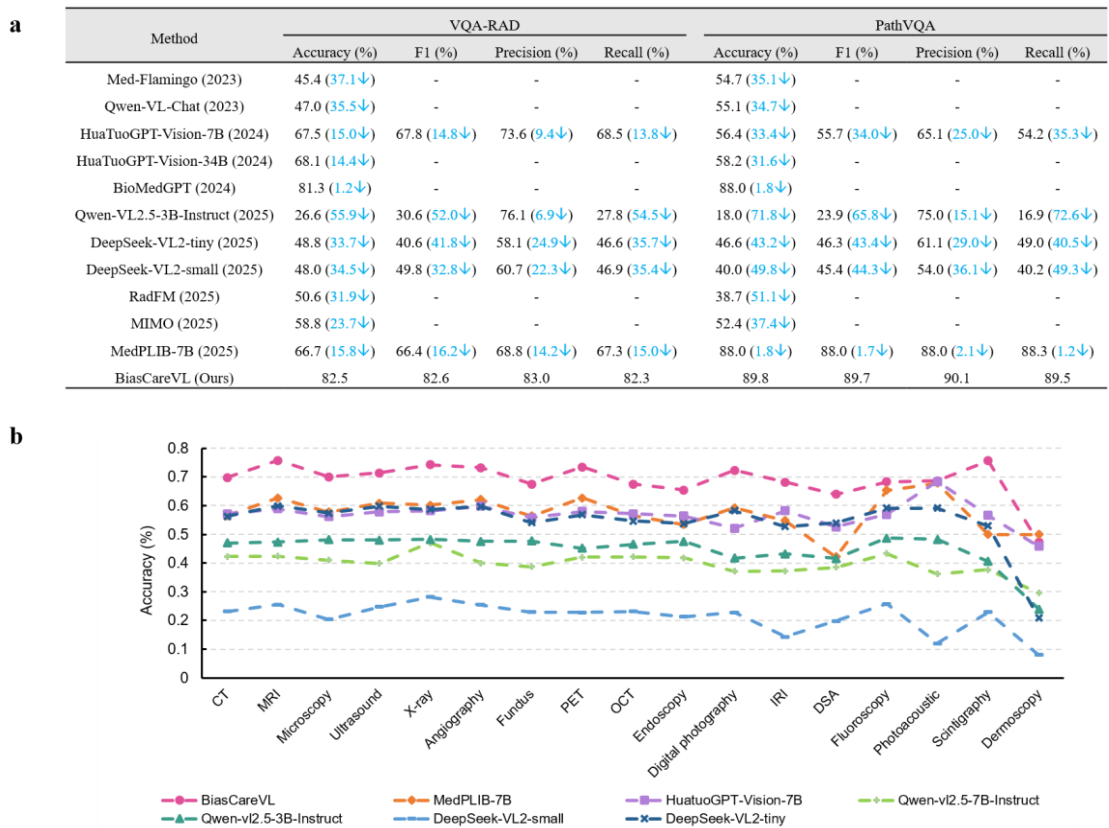

**a**,

| Method | VQA-RAD | | | | PathVQA | | | |
|---|---|---|---|---|---|---|---|---|
| | Accuracy (%) | F1 (%) | Precision (%) | Recall (%) | Accuracy (%) | F1 (%) | Precision (%) | Recall (%) |
| Med-Flamingo (2023) | 45.4 (37.1↓) | - | - | - | 54.7 (35.1↓) | - | - | - |
| Qwen-VL-Chat (2023) | 47.0 (35.5↓) | - | - | - | 55.1 (34.7↓) | - | - | - |
| HuaTuoGPT-Vision-7B (2024) | 67.5 (15.0↓) | 67.8 (14.8↓) | 73.6 (9.4↓) | 68.5 (13.8↓) | 56.4 (33.4↓) | 55.7 (34.0↓) | 65.1 (25.0↓) | 54.2 (35.3↓) |
| HuaTuoGPT-Vision-34B (2024) | 68.1 (14.4↓) | - | - | - | 58.2 (31.6↓) | - | - | - |
| BioMedGPT (2024) | 81.3 (1.2↓) | - | - | - | 88.0 (1.8↓) | - | - | - |
| Qwen-VL2.5-3B-Instruct (2025) | 26.6 (55.9↓) | 30.6 (52.0↓) | 76.1 (6.9↓) | 27.8 (54.5↓) | 18.0 (71.8↓) | 23.9 (65.8↓) | 75.0 (15.1↓) | 16.9 (72.6↓) |
| DeepSeek-VL2-tiny (2025) | 48.8 (33.7↓) | 40.6 (41.8↓) | 58.1 (24.9↓) | 46.6 (35.7↓) | 46.6 (43.2↓) | 46.3 (43.4↓) | 61.1 (29.0↓) | 49.0 (40.5↓) |
| DeepSeek-VL2-small (2025) | 48.0 (34.5↓) | 49.8 (32.8↓) | 60.7 (22.3↓) | 46.9 (35.4↓) | 40.0 (49.8↓) | 45.4 (44.3↓) | 54.0 (36.1↓) | 40.2 (49.3↓) |
| RadFM (2025) | 50.6 (31.9↓) | - | - | - | 38.7 (51.1↓) | - | - | - |
| MIMO (2025) | 58.8 (23.7↓) | - | - | - | 52.4 (37.4↓) | - | - | - |
| MedPLIB-7B (2025) | 66.7 (15.8↓) | 66.4 (16.2↓) | 68.8 (14.2↓) | 67.3 (15.0↓) | 88.0 (1.8↓) | 88.0 (1.7↓) | 88.0 (2.1↓) | 88.3 (1.2↓) |
| BiasCareVL (Ours) | 82.5 | 82.6 | 83.0 | 82.3 | 89.8 | 89.7 | 90.1 | 89.5 |

**Fig. 2 | Performance evaluation on VQA tasks. a**, Closed-ended VQA on the VQA-RAD and PathVQA benchmarks. BiasCareVL outperforms both general-purpose and medical vision-language models (VLMs). ↓ denotes the performance decrease compared to BiasCareVL. **b**, Closed-ended VQA across 16 imaging modalities in the

PMC-VQA benchmark, demonstrating BiasCareVL's robust performance on both common and rare modality types.

**Visual question answering across imaging modalities**

To assess the robustness of BiasCareVL against modality availability bias, we report its performance on closed-ended VQA tasks across 16 distinct medical imaging modalities from the PMC-VQA dataset[31] (Fig. 2b). The results demonstrate that BiasCareVL consistently outperforms representative baselines across the majority of modalities, showing strong reliability across both prevalent modalities and data-scarce modalities. Substantial gains are observed for underrepresented modalities. For example, on digital subtraction angiography (DSA) and elastography, accuracy improvements over the second-best method exceed 10%. Pronounced gains are also evident in modalities requiring complex cross-modal reasoning. In angiography and digital photography, BiasCareVL improves accuracy by 11.11% (73.19% vs. 62.08% for MedPLIP) and 13.99% (72.35% vs. 59.20% for MedPLIP), respectively. These improvements indicate that BiasCareVL's jointly learned vision-language embedding space successfully captures clinically meaningful semantic relationships, which generalize effectively across diverse imaging contexts. Together, these findings confirm that BiasCareVL not only achieves overall diagnostic superiority, but also maintains balanced and fair performance across a wide spectrum of data modalities. This represents a critical advance toward building reliable and generalizable medical vision-language intelligence for real-world clinical use.

**Multi-class skin disease classification under prevalence bias**

We assessed the quality of the visual representations learned by BiasCareVL by attaching a linear classification head to the vision embeddings and evaluated its performance on a long-tailed classification benchmark, ISIC2018[32,33]. The results were compared against several classical and state-of-the-art long-tailed learning approaches, including TFA-LT[34], ResLT[35], BCL[36], CLIP linear-prob[37], BALLAD[38], and Focal loss[39]. We emphasize that the core design of BiasCareVL is to perform disease diagnosis without attaching any task-specific head or fine-tuning, by directly leveraging the reasoning capability of the LLM. This experiment serves to demonstrate that the visual features learned by our framework are highly transferable and can be easily adapted to standard classification tasks with minimal adaptation. Following the protocol of TFA-LT[34], we divided the test set into three subgroups, many, medium, and few, based on class sample counts to systematically evaluate the model's robustness to disease prevalence bias. The results are summarized in Fig. 2a and Fig. 2b. Conventional baselines such as Focal loss exhibit severe performance degradation on the medium and few subgroups. Rebalancing-based methods like BALLAD and ResLT bring only marginal gains on the few subgroup, often at the cost of reduced accuracy on the many or medium subgroups. In contrast, multimodal methods such as BCL and TFA-LT deliver more balanced and competitive results. BiasCareVL outperforms all compared methods across all three subgroups, achieving an overall accuracy improvement of more than 10% and demonstrating superior capability in mitigating long-tailed data bias.

**Multi-label long-tailed disease classification**

To further investigate the capability of BiasCareVL on handling long-tailed data, we performed fine-tuning and testing on the CXR-LT2024 Task 2 benchmark[40]. Receiver operating characteristic (ROC) curves for tail and head classes are plotted in Fig. 3c. As expected, large differences in model performance across tail classes are observed. Among the four tail classes, the area under the curve (AUC) reaches 91.4% for the best-performing class, indicating strong discriminative ability. However, for certain diseases such as infiltration, the AUC drops substantially to 56.1%. A possible explanation is that infiltration exhibits imaging characteristics similar to those of other conditions (e.g., pneumonia), making it challenging for the model to learn stable representations. For the four head classes with larger sample sizes, the model performs more robustly, although variations across categories persist. Overall, the model achieves strong performance even on some rare categories.

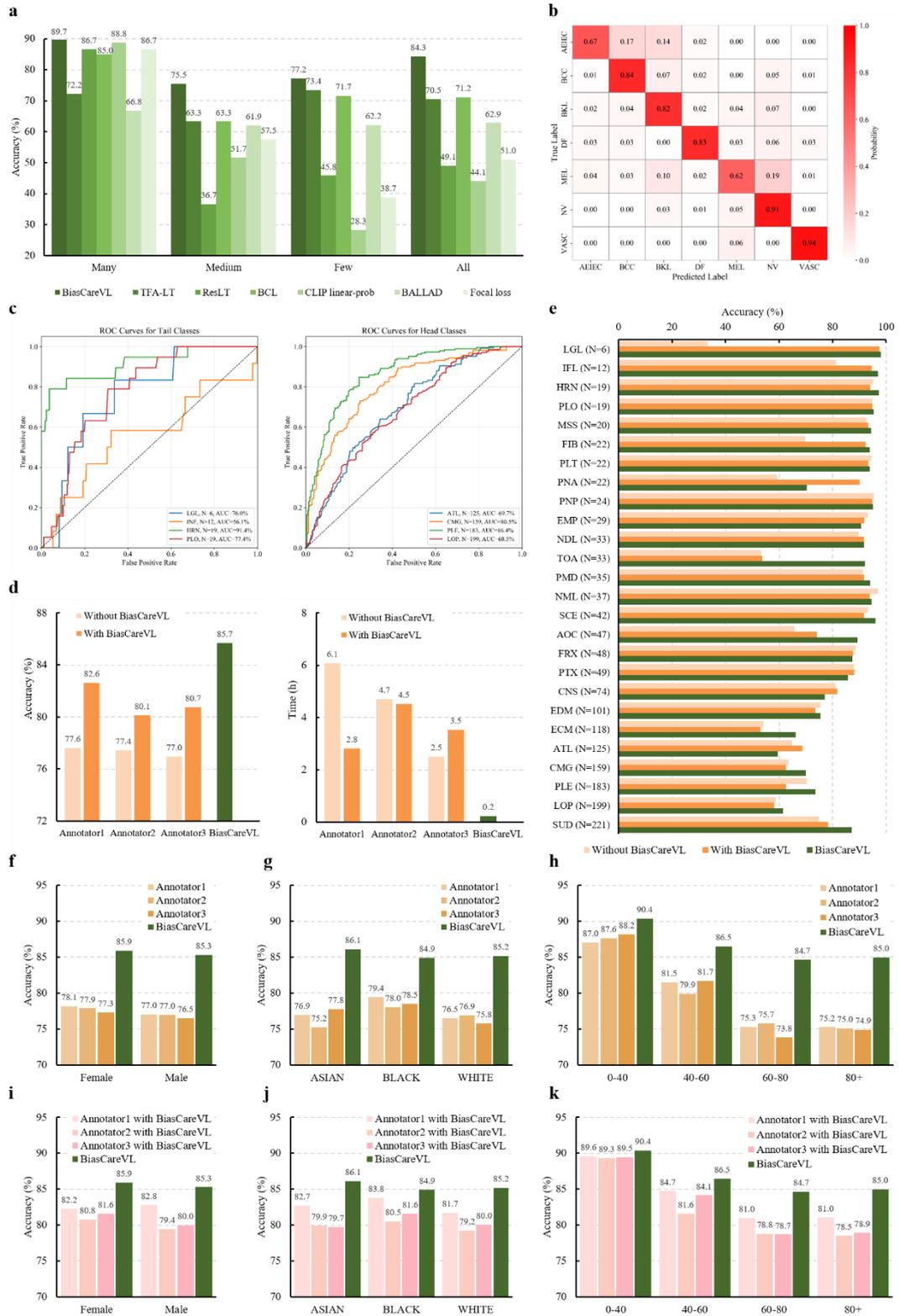

**Fig. 3 | Performance evaluation on classification tasks. a & b**, Skin disease classification on the ISIC2018 challenge. BiasCareVL achieves superior results across

all prevalence groups compared to existing long-tailed learning methods. **c**, ROC curves of tail and head classes of BiasCareVL on the CXR-LT 2024 dataset. **d,** Human expert and BiasCareVL performance for chest disease classification on the CXR-LT 2024 dataset. BiasCareVL surpasses experienced human annotators (>5 years) in accuracy while operating in substantially less time. While BiasCareVL assistance improves annotator performance, the annotation speed remains dependent on the individual annotator, suggesting that acceptance of AI predictions is human-dependent. **e**, Performance of Annotator 1 with and without BiasCareVL assistance, and performance of BiasCareVL itself, across 26 categories. **f-k**, Human expert and BiasCareVL performance for chest disease classification on the CXR-LT 2024 dataset across demographic subgroups (gender, race, and age). The 26 categories are: lung lesion (LGL), infiltration (INF), hernia (HRN), pleural other (PLO), mass (MSS), fibrosis (FIB), pleural thickening (PLT), pneumonia (PNA), pneumoperitoneum (PNP), emphysema (EMP), nodule (NDL), tortuous aorta (TOA), pneumomediastinum (PMD), normal (NML), subcutaneous emphysema (SCE), calcification of the aorta (AOC), fracture (FRX), pneumothorax (PTX), consolidation (CNS), edema (EDM), enlarged cardiomediastinum (ECM), atelectasis (ATL), cardiomegaly (CMG), pleural effusion (PLE), lung opacity (LOP), and support devices (SUD). N=6 indicates that in the test set, there are six data classified as the corresponding category.

**Disease classification performance compared to human experts**

We conducted a human-AI comparison on the CXR-LT2024 Task 2 benchmark[40]. Three experts were provided with the images and the 26 disease categories and were asked to independently classify each image into the appropriate categories. To facilitate the annotation process, we built a simple graphical user interface that automatically presents the images and saves the results; the experts only need to select the categories, eliminating time spent on image selection or saving results. Annotation was performed in two rounds: without and with BiasCareVL assistance. Without AI assistance, all three experts achieved similar accuracy, with Annotator 1 taking the longest time (6.1 hours). BiasCareVL achieved better accuracy than all three annotators, outperforming them by 8.1% in absolute accuracy with much shorter time, speeding up the process by more than 10 times. When assisted by BiasCareVL, all three annotators showed improved annotation performance. Notably, Annotator 1's accuracy increased by 5.0% while the time required dropped to 2.8 hours. A category-specific analysis reveals that the largest performance gains occurred for rare cases, such as lung lesion (LGL; accuracy increased by 64.3%) and fibrosis (FIB; accuracy increased by 22.6%), highlighting the promise of AI-assisted annotation in tail cases.

**Disease classification performance across demographic subgroups**

To assess whether BiasCareVL maintains equitable performance across different populations, we stratified the results on the CXR-LT2024 Task 2 dataset by gender, race, and age. For each subgroup we computed per-class accuracy, then averaged them macro-wise over the 26 disease categories (Fig. 3f-k). Both the human experts and BiasCareVL achieved similar accuracy for female and male patients, with a difference smaller than 1.6%. Across the three racial groups (Asian, Black, and White), BiasCareVL's macro-averaged accuracy ranged from 84.9% to 86.1%, a maximum difference of 1.2%. Human experts without AI assistance showed a larger spread across races (2.7%–2.9% for the three annotators). For the four age groups, we observed declining performance in the two older cohorts. BiasCareVL outperformed human experts in every age bracket, particularly for patients aged 80 and above. The maximum difference across age groups was 5.7% for BiasCareVL, much smaller than those for the unaided human experts, which were 11.8%, 12.6%, and 14.4% for the three annotators, respectively. When assisted by BiasCareVL, these performance variations were reduced, falling to 8.5%, 10.8%, and 10.8%.

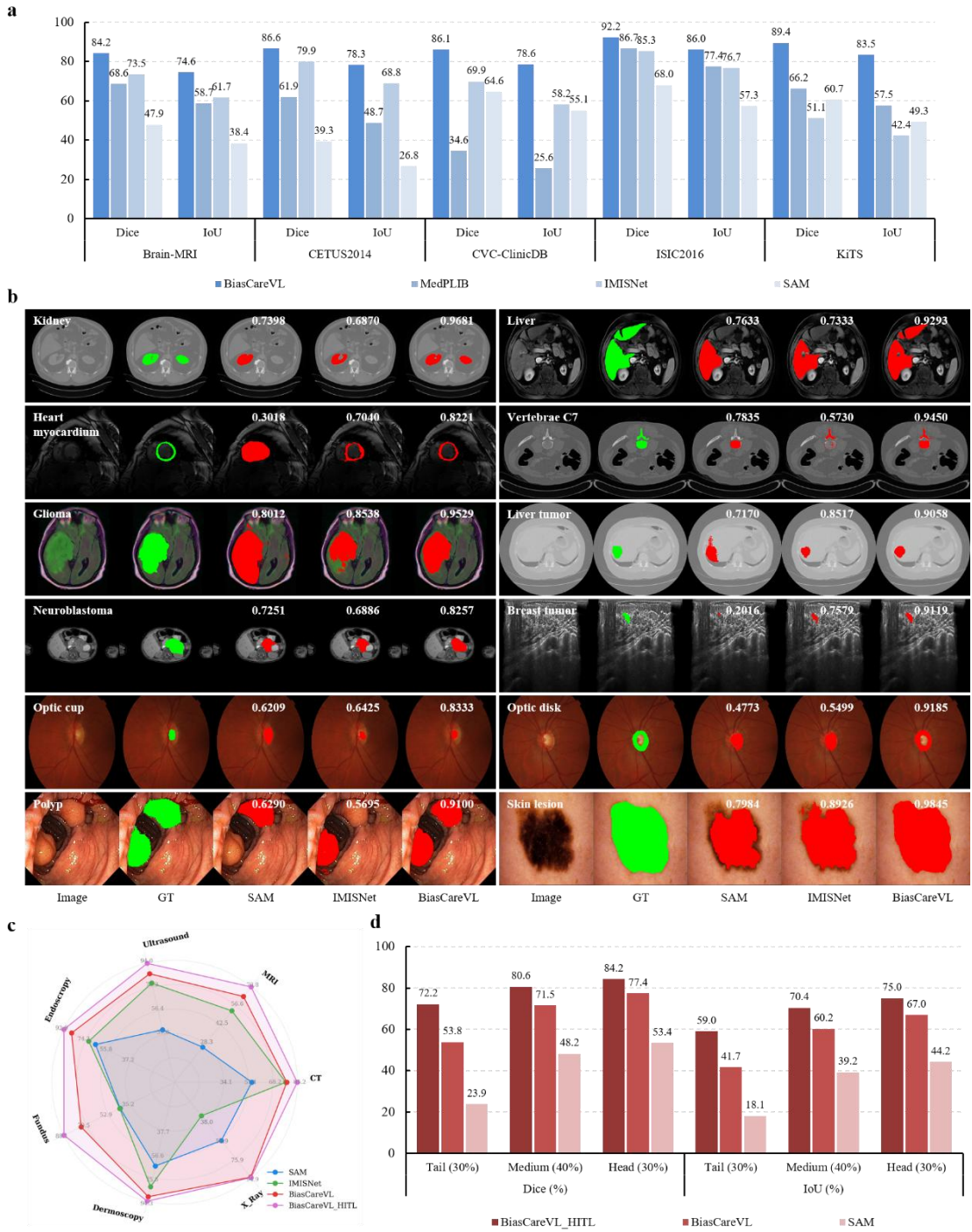

**Fig. 4 | Performance evaluation on segmentation tasks. a**, Segmentation results on five randomly selected datasets from IMed-361M. BiasCareVL outperforms both general-purpose and medical vision-language models (VLMs). **b**, Visual comparison of segmentation masks across diverse scenarios. Dice scores are annotated at the top-right

corner. **c**, Performance breakdown (Dice) across different imaging modalities. BiasCareVL_HITL denotes simulation of human-in-the-loop inference using a single point correction. **d**, Performance breakdown (Dice and IoU) for data subgroups with varying category prevalence levels.

**Organ and lesion segmentation across diverse public datasets**

To evaluate the organ-at-risk and pathology localization performance of BiasCareVL, we conducted segmentation experiments on the IMed-361M dataset[41], which is a recently released large-scale medical image segmentation benchmark that aggregates 110 publicly available datasets. In Fig. 4a we report results on five selected datasets of different modalities: Brain-MRI (brain tumor segmentation in MRI), CETUS2014 (left ventricle segmentation in ultrasound), CVC-ClinicDB (polyp segmentation in endoscopy), ISIC2016_Task1 (skin lesion segmentation in dermoscopy), and KiTS (kidney tumor segmentation in CT). We compare against three representative models: SAM-ViT-B[42], IMISNet-B[41], and MedPLIB-7B[30]. As shown in Fig. 4a, BiasCareVL achieves the best performance on all five datasets. On the five selected datasets, it not only improves results on relatively well-represented tasks such as skin lesion segmentation, but also delivers substantially higher scores on challenging 3D tumor segmentation tasks, exceeding the Dice scores of the strongest baseline by more than 23% on kidney tumors and 15% on brain tumors. Fig. 4b visualizes typical segmentation outputs across datasets with widely varying imaging modalities and target sizes. BiasCareVL produces more accurate boundaries that align closely with ground-truth masks in all scenarios, consistently outperforming the comparison methods. Together, these results demonstrate BiasCareVL's strong generalization capability across diverse anatomical structures and imaging modalities, spanning from 2D dermoscopic images to 3D MRI volumes and from small targets (e.g., kidney tumors) to large ones (e.g., skin lesion).

**Segmentation performance under category and modality biases**

To further assess the robustness of BiasCareVL's segmentation, we analyze its performance across subgroups defined by imaging modality (Fig. 4c) and category prevalence (Fig. 4d). While the IMed-361M dataset originally covers 14 imaging modalities, we consolidate different MRI sequences (e.g., T1-weighted, T2-weigtehd) into a single "MRI" category, resulting in seven distinct modality groups: MRI, CT, X_ray, demoscropy, fundus, endoscropy, and ultrasound. As shown in Fig. 4c, BiasCareVL achieves the most balanced and robust segmentation results across all seven modalities compared with SAM and IMISNet, demonstrating its ability to mitigate modality availability bias. Similarly, when evaluated on subgroups of varying prevalence (head, medium, tail), BiasCareVL substantially outperforms the SAM baseline across all three groups (Fig. 4d). The improvement is particularly pronounced for the tail (low-prevalence) subgroup, where the Dice score increases by 29.9%, compared to a 24.0% gain for the head subgroup, highlighting the model's effectiveness in handling anatomical and disease prevalence bias. We also report results with one-point expert feedback during inference (denoted as BiasCareVL_HITL in Fig. 4c and Fig. 4d). When a single corrective point prompt is provided via the HITL module, performance is consistently further improved. This illustrates the flexibility of BiasCareVL in real clinical workflows, where optional expert input can be efficiently employed to verify and refine model predictions.

**High-quality and coherent medical report generation**

To evaluate the medical report generation capability of BiasCareVL, we conducted experiments on two large-scale benchmarks, MIMIC[43] and PubMedVision[44]. We compared against seven state-of-the-art models: LLaVA-Med v1.5[45], HuatuoGPT-Vision-7B[44], MedPLIB-7B[30], DeepSeek-VL2-tiny[28], DeepSeek-VL2-small[28], Qwen-VL2.5-3B-Instruct[27], and Qwen-VL2.5-7B-Instruct[27].

Overall, BiasCareVL achieves superior report generation performance on both benchmarks (Fig. 5). On the widely adopted MIMIC dataset, BiasCareVL improves the BLEU-1 score by 52.14% (from 26.64 for DeepSeek-VL2-small to 40.53) and the ROUGE-1 score by 44.46% (from 22.38 for HuaTuoGPT-Vision-7B to 32.33) relative to the respective best baselines. Moreover, BiasCareVL generates consistent and balanced results across the two benchmarks, whereas the comparison models often exhibit biased performance. For example, the scores of HuaTuoGPT-Vision-7B on PubMedVision are nearly twice as high as those on MIMIC, while BiasCareVL maintains closely matched scores across both datasets. Furthermore, while DeepSeek-VL2-small attains a slightly higher BLEU-1 score than BiasCareVL on PubMedVision, its ROUGE-1 score on the same benchmark is substantially lower, revealing a metric-level inconsistency that our model does not exhibit. These results underscore the robustness and reliability of BiasCareVL for clinical report generation, demonstrating its ability to produce high-quality, semantically coherent textual descriptions across diverse medical imaging contexts.

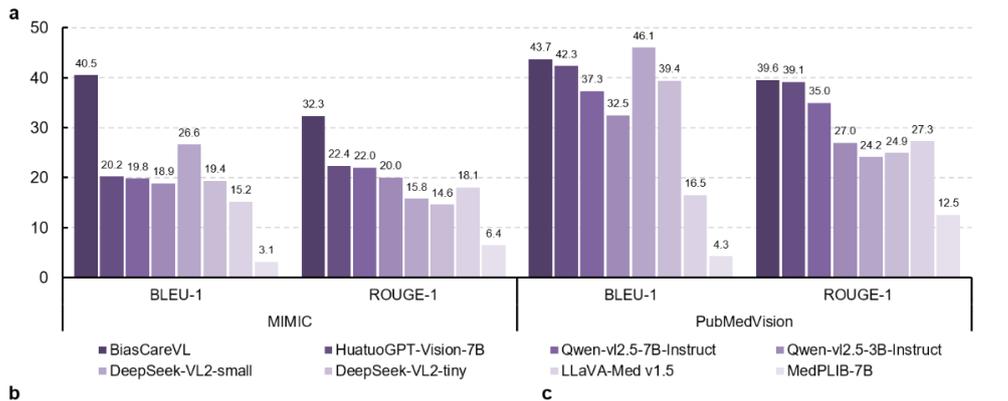
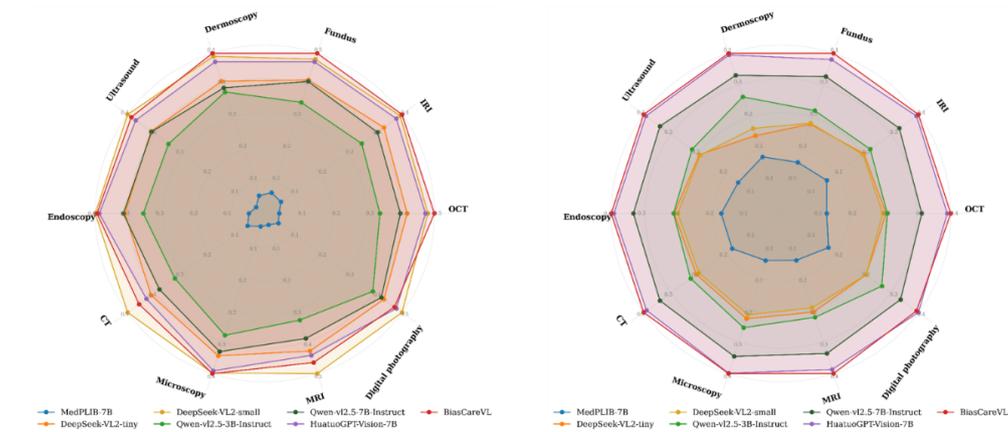
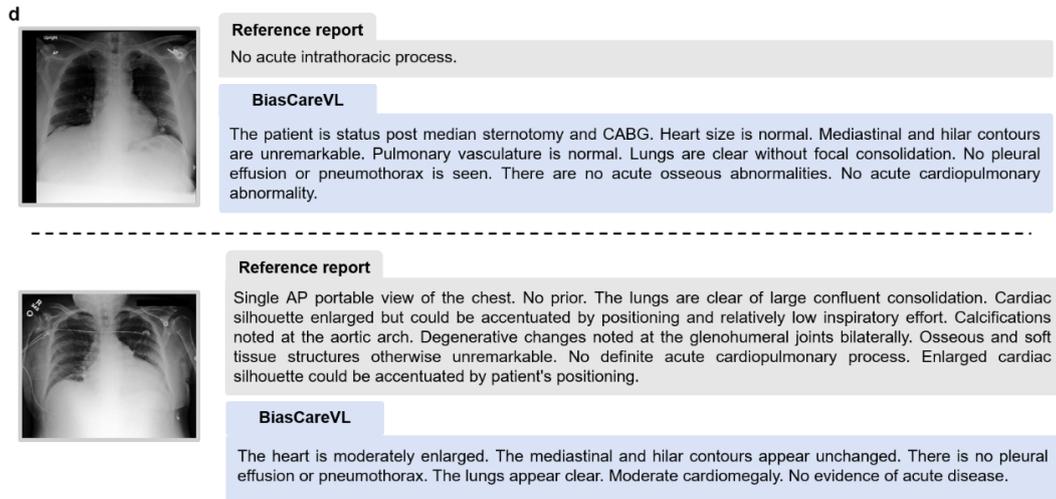

**Fig. 5 | Performance evaluation on report generation tasks. a**, Report generation performance (BLEU-1 and ROUGE-1 scores) on the MIMIC and PubMedVision benchmarks. **b & c**, Report generation performance, BLEU-1 and ROUGE-1 scores, across ten imaging modalities on the PubMedVision benchmark. **d**, Two example reports generated by BiasCareVL.

**Robust report generation across imaging modalities**

To further assess the robustness of BiasCareVL for report generation across different imaging modalities, we analyze its modality-wise performance on the PubMedVision dataset (Fig. 5b and Fig. 5c). BiasCareVL generates the best performance on most modalities considering both BLEU-1 and ROUGE-1, confirming its robustness to modality availability bias. Comparing BiasCareVL with DeepSeek-VL2-small, which generates a slightly higher overall BLEU-1 score (46.05) than BiasCareVL (43.74), we observe that the BLEU-1 differences between the two models are marginally small ($-1.31\% \pm 4.39\%$, ranging from -8.30% on CT to 3.96% on OCT). In contrast, the ROUGE-1 differences are substantial: BiasCareVL outperforms DeepSeek-VL2-small by $68.83\% \pm 8.79\%$ (from 57.93% on microscopy to 88.30% on demoscropy). Since BLUE-1 primarily reflects lexical fidelity (word-level overlap) while ROUGE-1 better captures content relevance and semantic coverage, the pronounced ROUGE-1 advantage indicates that BiasCareVL generates more semantically accurate and clinically relevant reports, even when lexical surface similarity is comparable. This result underscores the model's ability to produce high-quality textual descriptions that are robust across diverse imaging modalities. The generated report examples further demonstrate that BiasCareVL consistently provides clinically correct information (Fig. 5d).

**Discussion**

In this study, we present BiasCareVL, a fully open medical vision-language framework designed to care for biased data distributions, thereby advancing fair, robust, and generalizable multimodal clinical intelligence. BiasCareVL addresses a central challenge in clinical artificial intelligence: the pervasive and inherent biases in medical data[5,6], with a focused mitigation of disease prevalence bias, modality availability bias, and demographic bias[5,17]. Through benchmarking on eight large-scale public datasets spanning diverse tasks, including closed-ended VQA, multi-class and multi-label disease classification, organ-at-risk and pathology segmentation, and report generation, we demonstrate that BiasCareVL achieves superior performance compared to existing state-of-the-art methods. Additionally, we compare BiasCareVL with human experts on multi-label disease classification, showing that it achieves higher accuracy in substantially less time.

While biased data distributions represent a long-standing issue in medical data analysis, with established mitigation strategies such as data augmentation, reweighting, and ensemble learning[20,21], this challenge remains relatively underexplored in the context of medical VLMs[11,29]. BiasCareVL confronts this gap by introducing adaptive uncertainty modeling with an optional HITL refinement mechanism. This designed method effectively reduces performance degradation on skewed datasets by dynamically quantifying and responding to prediction unreliability. The uncertainty estimation, derived from both semantic misalignment and spatial error, serves as an internal signal to prioritize challenging and underrepresented samples during training. Concurrently, the simulated feedback loop allows the model to learn from its own segmentation or diagnostic errors in a self-corrective manner.

To examine whether BiasCareVL yields equitable performance across different subgroups, we report results across disease groups, imaging modalities, and population groups. For disease prevalence bias, BiasCareVL achieves higher accuracy across all subgroups in multi-class skin disease classification (13.1% improvement over the second-best method) and more balanced results across the 26 categories in multi-label chest disease classification (Fig. 3d). Similarly, on segmentation tasks, BiasCareVL delivers superior performance for both common and rare targets (Fig. 4a, b, d), increasing the Dice score for small kidney segmentation by 23.2%. Regarding imaging modalities, BiasCareVL maintains more balanced performance across modalities than existing state-of-the-art methods on closed-ended VQA (Fig. 2b), segmentation (Fig. 4c), and report generation tasks (Fig. 5b, c). For demographic bias, we compare human experts and BiasCareVL across gender, race, and age. BiasCareVL exhibits smaller accuracy variations across these subgroups than all three unaided experts, confirming its improved fairness across population groups.

A distinctive strength of BiasCareVL lies in its operational flexible, enabled primarily by the optional HITL module during inference. This feature allows clinicians to review, and if necessary, refine uncertain predictions with minimal effort, for example, by adding corrective click prompts. Such interactive capability not only elevates segmentation and diagnostic accuracy in ambiguous cases but also fosters greater clinician trust and acceptance of AI assistance. By aligning with real-world clinical workflows where expert oversight is paramount, BiasCareVL transitions from a black-box automated tool into a collaborative decision-support system, paving the way for smoother integration into routine practice.

BiasCareVL is designed as a general and extensible framework rather than a fixed model. Its dual-path architecture permits the MLLM pathway to be initialized with various foundation models (e.g., Qwen-VL[26,27], LLaVA-Med[45], HuatuoGPT-Vision[44]) and the vision pathway to incorporate evolving segmentation encoders (e.g., IMIS-Net[41] or future variants). This modularity ensures that the framework can continuously integrate advances in both language and vision domains, maintaining relevance as newer and more capable base models emerge.

While BiasCareVL represents a significant advancement in bias-aware multimodal AI for clinical decision-making, there are areas where further development could enhance its capabilities. First, despite BiasCareVL has been evaluated over 15 imaging modalities and hundreds of thousands of multimodal samples, certain ultra-rare conditions remain sparsely represented, and their generalization may benefit from privacy-preserving federated learning across institutions[46] or data generation using advanced generative models[47–49]. Second, while the adaptive uncertainty modeling mechanism successfully prioritizes uncertain samples, future versions could benefit from incorporating causal or mechanistic understanding of disease processes.

Integrating structured medical knowledge graphs or pathology-aware constraints may provide an additional layer of robustness, especially in highly atypical cases, and improve interpretabilit[50]. Looking ahead, future work will focus on expanding BiasCareVL's coverage of rare diseases, exploring scalable self-supervised learning under extreme imbalance, and integrating richer forms of clinical knowledge into unified multimodal representations. Further development of interactive interfaces may also enhance the synergy between clinicians and AI, enabling rapid refinement, robust auditing, and real-time decision support.

In summary, BiasCareVL represents an important step toward building robust, fair, and generalizable medical vision-language intelligence. By directly addressing data bias through uncertainty-guided learning and optional human collaboration, it achieves superior and balanced performance across a wide range of tasks and data regimes. We open-source the complete framework to promote transparency, reproducibility, and collaborative advancement in the field. It is our hope that BiasCareVL will serve as a foundational tool for developing the next generation of trustworthy multimodal AI systems in medicine.

**Methods**

**Model architecture**

The overall framework of BiasCareVL is illustrated in Fig. 1g. BiasCareVL consists of four key components: a SAM-based vision pathway that extracts anatomical and pathological representations from medical images, a large-scale multimodal LLM (MLLM) that performs semantic reasoning and cross-modal alignment, an adaptive uncertainty modeling module that adaptively leverages clinical expertise for uncertain or underrepresented cases, and an option human-in-the-loop module that identifies difficult cases for expert review. BiasCareVL unifies vision and language processing within a single architecture, enabling it to support a wide variety of clinical tasks, including VQA, report generation, disease classification, and image segmentation, without requiring task-specific model design or fine-tuning unless otherwise specified.

Given an input image $I \in \mathbb{R}^{3 \times H \times W}$, a text prompt $Q_t$, and a vision prompt $Q_v$, BiasCareVL processes these inputs through two parallel but interconnected pathways: (1) The MLLM pathway is optimized for language-conditioned tasks such as VQA and report generation. It encodes the image $I$ and text prompt $Q_t$ into corresponding embeddings $z_v \in \mathbb{R}^{l1 \times d}$ and $z_t \in \mathbb{R}^{l2 \times d}$, where $l1$ and $l2$ are the number of visual and text tokens and $d$ is the embedding dimension. These two sets of embeddings are concatenated and input into a large language model (LLM) backbone. The LLM processes the sequence to generate a contextualized sequence of hidden representations $z_{vlm} \in \mathbb{R}^{L \times d}$, where $L = l1 + l2$. These hidden representations capture rich semantic information for downstream multimodal reasoning. (2) The vision pathway handles image-based tasks such as classification and segmentation. A SAM-based vision encoder extracts dense visual embeddings $z_{img} \in \mathbb{R}^{C \times h \times w}$, which encodes anatomical structure and pathological patterns in high resolution.

To bridge the two modalities, a cross-modal feature alignment module integrates semantic information from the MLLM into the vision pathway. Specifically, a set of learnable query vectors $z_q \in \mathbb{R}^d$ is introduced to attend to the MLLM's hidden states $z_{vlm}$ via multi-head attention, producing pooled semantic embeddings:

$$\hat{z}_{vlm} = MultiheadAtt(z_q, z_{vlm}, z_{vlm}) \tag{1}$$

The pooled semantic features $\hat{z}_{vlm} \in \mathbb{R}^d$ are then projected through a linear transformation:

$$\bar{z}_{vlm} = W_p \cdot \hat{z}_{vlm} + b_p \tag{2}$$

Finally, $\bar{z}_{vlm} \in \mathbb{R}^C$ is concatenated with the visual prompt embedding $z_{vp} \in \mathbb{R}^{N \times C}$ ($N$ represents the number of points when point prompt is provided or $N = 2$ when box prompt is provided) to form a semantically enhanced visual prompt, which is used to guide the vision decoder. This enables the model to generate more accurate and context-aware segmentation outputs. By unifying the MLLM and vision pathways, the model supports both coarse-grained language tasks and fine-grained visual tasks within a single framework.

**Bias mitigation via adaptive uncertainty modeling and human-in-the-loop refinement**

To address the performance degradation caused by bias in medical datasets (e.g., modality availability bias in Figure 1(a) and anatomical or disease prevalence bias in Figure 1(b)), we introduce two complementary components: adaptive uncertainty-guided learning and optional human-in-the-loop (HITL) refinement.

**Adaptive uncertainty modeling.** An adaptive uncertainty modeling module is proposed that quantifies prediction reliability at both semantic and spatial levels. This uncertainty signal is not only used to guide learning during training but also serves as a criterion for expert review in the HITL module.

We first measure the semantic alignment between the prediction of the mask decoder and the corresponding textual concept to evaluate semantic consistency-based uncertainty. The dense visual embeddings $z_{img}$ from the vision encoder are upsampled via bilinear interpolation to generate $Z_{img} \in \mathbb{R}^{C \times H \times W}$. A mask-weighted global visual embedding $\bar{z}_{img} \in \mathbb{R}^C$ is then computed as:

$$\bar{z}_{img} = \frac{\sum_{x,y}(Z_{img}(x,y) \cdot P(x,y))}{\sum_{x,y} P(x,y) + \varepsilon_1} \tag{3}$$

where $P(x,y)$ denotes the confidence map predicted by the mask decoder, and $\varepsilon_1 = 1e^{-6}$ ensures numerical stability. This formulation emphasizes feature activations in high-confidence regions while suppressing background responses.

Next, we compute the cross-modal similarity between the visual embedding $\bar{z}_{img}$ and the pooled language embedding $\bar{z}_{vlm}$ using cosine similarity:

$$s_{vl} = \frac{\langle \bar{z}_{img}, \bar{z}_{vlm} \rangle}{\|\bar{z}_{img}\|_2 \cdot \|\bar{z}_{vlm}\|_2} \tag{4}$$

where $\langle , \rangle$ indicates the cosine similarity operation, and $\|\cdot\|_2$ calculates the $L_2$ norm. A smaller $s_{vl}$ indicates weaker consistency between visual evidence and linguistic semantics. The corresponding semantic consistency-based uncertainty is thus defined

as:

$$u_{vl} = 1 - s_{vl} \tag{5}$$

To capture both semantic and spatial unreliability, the final uncertainty score is computed by combining the semantic term $u_{vl}$ with the Dice loss $l_{dice}$, which quantifies pixel-level segmentation error:

$$u = \beta_{vl} \cdot u_{vl} + l_{dice} \tag{6}$$

where $\beta_{vl} = 0.5$ is a balancing coefficient controlling the relative contribution of vision-language inconsistency. A larger $u$ reflects greater unreliability due to semantic misalignment, spatial error, or both.

To emphasize uncertain and underrepresented samples during model training, the joint uncertainty $u$ is transformed into a dynamic sample weight via an exponential transformation:

$$w = \exp(\lambda_u \cdot u) \tag{7}$$

where $\lambda_u = 1.0$ acts as a temperature parameter, controlling the sensitivity of the weighting mechanism. The resulting weight $w$ is then used to re-scale the Dice loss $l_{dice}$. This mechanism adaptively assigns higher learning priority to uncertain samples by amplifying their loss contributions, encouraging the model to focus on correcting challenging or semantically inconsistent predictions. The joint uncertainty modeling serves as an adaptive curriculum, focusing the model's capacity on semantically ambiguous or spatially challenging samples. This leads to more stable training dynamics and fosters the learning of bias-invariant representations, which directly improves generalization across heterogeneous and biased medical datasets.

**Human-in-the-loop refinement.** To further enhance the robustness of BiasCareVL in high-stakes or ambiguous scenarios, we incorporate an optional HITL module. This design mirrors the clinical workflow where radiologists review and refine uncertain preliminary findings, enabling dynamic error correction and self-improvement through iterative feedback. The process operates in two distinct modes: a self-refinement mode during training, which simulates expert feedback using ground truth, and a true interactive mode during inference, which incorporates actual expert input.

In the training phase, the model identifies and prioritizes its own mistakes for focused learning, using ground truth as an oracle to simulate expert correction. Given a batch of input images $I_B \in \mathbb{R}^{B \times 3 \times H \times W}$, we compute the corresponding uncertainty scores $u_B \in \mathbb{R}^B$ using Eq. (6). The top $k$ samples ($k = r \cdot B$, $r = 0.3$) with the highest uncertainty, denoted set $\mathcal{H}$, are selected. These represent cases with poor semantic alignment or high segmentation error.

For each hard example $I_i \in \mathcal{H}$, the model generates a binary mask prediction $M_i$. Instead of requiring a human expert, we simulate the refinement step by comparing $M_i$ with the ground-truth mask $Y_i$. Specifically, the error region $E_i$ is obtained via the logical XOR operation:

$$E_i = M_i \oplus Y_i \qquad (8)$$

A set of corrective prompt points is sampled from $E_i$ and encoded by the vision prompt encoder. Combined with the original visual prompt embeddings $z_{vp}$, refined prompt embeddings $\hat{z}_{vp}$ are generated for a subsequent forward pass. This procedure forces the model to concentrate its learning capacity on confusing regions, effectively implementing a form of self-paced, error-driven curriculum learning.

During the inference phase, BiasCareVL operates in an interactive mode designed for seamless integration into clinical workflows. Particularly, for a given input, the model produces an initial segmentation mask alongside its uncertainty estimate, calculated via Eq. (5). The expert can choose to accept the initial segmentation or initiate a correction. If refinement is needed, the expert can efficiently correct errors using intuitive interactions (e.g., adding positive/negative clicks). These corrective prompts are fed back into the model in real-time, generating a refined segmentation.

This dual-phase HITL design creates a natural pathway for a clinically viable semi-automated annotation system. It leverages human expertise efficiently, directing it only to the most uncertain and high-value cases, while automating routine, confident predictions. Consequently, the system not only improves its own robustness through iterative self-correction during training but also guarantees reliable, expert-verified outcomes in critical real-world applications.

**Multi-task optimization objectives**

BiasCareVL is optimized through a unified objective that jointly coordinates the vision and vision-language pathways, reflecting a principled integration of complementary learning signals. The overall loss function integrates three objectives, each designed to address a specific aspect of robust model learning.

**Adaptive uncertainty-weighted segmentation loss.** The primary segmentation loss leverages the estimated joint uncertainty to dynamically balance sample importance during training. This loss is computed over all samples in a training batch:

$$L_{AUR} = \sum_{i \in I_B} v_i \cdot w_i \cdot l_{dice,i} \quad (9)$$

where $v_i \in \{0, 1\}$ is a binary indicator for the presence of a valid segmentation label. $l_{dice,i}$ is the standard Dice loss for sample $i$. $w_i$ is the dynamic uncertainty weight defined in Eq. (7). This formulation prioritizes challenging samples with high uncertainty, whether stemming from semantic inconsistency or spatial error, effectively implementing a self-paced curriculum that focuses model capacity on the most difficult regions of the data distribution.

**Human-in-the-loop refinement loss.** To explicitly correct high-uncertainty errors, an additional loss is applied exclusively to the subset of hard samples $\mathcal{H}$ selected for simulated HITL refinement:

$$L_{HITL} = \sum_{i \in \mathcal{H}} v_i \cdot l_{refine,i} \quad (10)$$

where $l_{refine,i}$ is the Dice loss calculated between the ground truth and the refined prediction obtained after processing simulated expert feedback through the vision prompt encoder and mask decoder. This loss closes the feedback loop, enabling the model to learn from its own mistakes and directly optimize for iterative improvement akin to expert-guided correction.

**Vision-language alignment loss.** The MLLM pathway is optimized using a cross-entropy loss over next-token prediction, facilitating natural language reasoning and report generation:

$$L_{VLM} = \sum_t \log P(t_k | t_{<k}, z_v, z_t; \Theta_{vlm}) \tag{11}$$

where $t_k$ is the target token at position $k$. $t_{<k}$ denotes all receding tokens. $z_v$ and $z_t$ are the input sequence of LLM. $\Theta_{vlm}$ represents the trainable parameters of the MLLM pathway. This objective fosters semantically consistent multimodal reasoning by aligning visual representations with clinical language.

The complete training objective integrates these three losses into a unified formulation:

$$L = L_{AUR} + L_{HITL} + L_{VLM} \tag{12}$$

This unified optimization design reflects clinical decision-making principles: uncertain or atypical cases receive greater weighting, difficult samples are refined through simulated human feedback, and vision-language alignment enables interpretable reasoning. This integrated approach results in a model that is both technically robust and clinically reliable, demonstrating strong generalization across biased medical datasets.

**Implementation details**

All experiments were conducted on a server equipped with four NVIDIA A800 GPUs (80 GB memory each). We implemented our framework using DeepSpeed to enable efficient large-scale model training[51]. Training was performed with the AdamW optimizer using a base learning rate of $1 \times 10^{-5}$, while the MLLM pathway was assigned a lower learning rate of $5 \times 10^{-6}$ to stabilize multimodal alignment. The batch size was set to 4. To mitigate catastrophic forgetting and reduce computational overhead, we integrated Low-Rank Adaptation (LoRA) via the Parameter-Efficient Fine-Tuning (PEFT) framework[52], with a LoRA rank of 8 and scaling hyperparameter α = 16. All input images were uniformly resized to $256 \times 256$ pixels.

We initialized the MLLM pathway with HuatuoGPT-Vision[44] and the vision pathway with IMIS-Net[41]. Model training was carried out in two stages. In the first stage, we trained the vision pathway solely on the IMed-361M dataset for one day to adapt the model to the target input resolution. In the second stage, we incorporated the MLLM pathway, and the entire model was jointly trained the model on a mixture of the IMIS-Net dataset and multiple VQA datasets. A 1:1 sampling ratio was maintained between the segmentation and VQA samples to balance the two learning objectives. This stage required approximately 25 days of continuous training under the described hardware setup. This training strategy ensures that both pathways are sufficiently adapted to the medical domain while preserving the semantic reasoning capabilities of the pre-trained MLLM backbone.

**Comparison methods**

We compared BiasCareVL with several state-of-the-art methods encompassing both vision and vision-language tasks. For visual segmentation, we included SAM[42] and IMIS-Net[41]. IMIS-Net is a SAM-based model pretrained on large-scale medical image datasets, representing a strong medical image segmentation baseline. For multimodal benchmarks, we considered DeepSeek-VL2[28], Qwen-VL-2.5[27], HuatuoGPT[44], BioMedGPT[3], MIMO[53], and MedPLIB[30]. To ensure a fair comparison with SAM-based methods that require at least one point prompt during inference, we provided a single randomly sampled point prompt for all segmentation evaluations unless otherwise specified. Within our GPU memory constraints, we adopt the largest feasible variant of each model series to ensure a representative and competitive evaluation. Detailed configurations of the compared models are reported in the corresponding result tables. All baseline models were re-implemented within our unified framework to guarantee consistent evaluation protocols and to eliminate implementation-specific biases.

**Evaluation metrics**

We employ a comprehensive set of widely adopted evaluation metrics to assess model performance across different tasks. For language-related tasks, including visual question answering (VQA), clinical report generation, and lesion/structure detection, we follow the established practices of prior works[27,44]. Specifically, for closed-ended VQA, we report accuracy as the primary metric, reflecting the discrete, correctness-based nature of the task. For open-ended VQA, we adopt recall, which better captures the model's ability to retrieve clinically relevant answers in scenarios where multiple valid responses may exist. For report generation, we use BLEU-1 and ROUGE-1 scores to evaluate both the lexical fidelity and the content relevance of generated reports against reference reports, balancing surface-form overlap with semantic relevance. For segmentation tasks, we report the Dice coefficient and Intersection-over-Union (IoU) to quantify pixel-level accuracy. The Dice coefficient emphasizes overlap similarity and is particularly sensitive to small or irregular structures, while IoU provides a stricter measure of spatial agreement. Together, these two metrics offer a balanced assessment of segmentation performance across both common and rare anatomical regions.

**Datasets**

To assess the performance of the proposed method, we conduct experiments on several widely used public datasets with sever biased distributions in imaging modality availability and disease/anatomic prevalence (Fig. 1a, b, and c). In the following, we briefly summarize their main characteristics, including dataset composition and data splitting strategies.

**PubMedVision**[44] is a large-scale medical VQA dataset created by extracting high-quality image-text pairs from PubMed articles and refining them using GPT-4V. It consists of two subsets. For our experiments, we utilize the Alignment subset, which contains 647,031 image-text pairs spanning diverse imaging modalities. Since no official division is provided for report generation, we randomly select 0.25% of the data (1,617 samples) as the test set, with the remainder used for training.

**IMed-361M**[41] aggregates 110 publicly available medical segmentation datasets, forming one of the largest repositories of medical segmentation masks to date. Spanning 14 imaging modalities, it comprises over 361 million annotated masks. The dataset covers 195 distinct anatomical structures and disease categories, including 1,983,539 labeled instances for training and 209,353 for testing. This extensive collection enables robust evaluation under clinically realistic distribution shifts.

**MIMIC-CXR v2**[43] is a large-scale chest X-ray database comprising 377,110 radiographs paired with 227,827 corresponding clinical reports. It has become a widely adopted benchmark for radiology-related research tasks. Following the official release, there are 368,960 training samples, 2,991 validation samples, and 5,159 testing samples.

**PMC-VQA**[31] comprises 227k question–answer pairs associated with 149k medical images, covering diverse imaging modalities and disease categories. We employ the official version-2 split, which consists of 152,603 training and 33,430 validation samples. The dataset is structured in a multiple-choice format and serves to evaluate medical closed-ended VQA performance.

**VQA-RAD**[24] focuses on radiology images and is designed to benchmark medical VQA systems. It includes both open-ended and binary yes/no questions. We adopt the official split, consisting of 1,793 training and 451 testing question-answer pairs.

**PathVQA**[25] contains 4,998 pathology images and 32,799 associated question-answer pairs. Among them, 4,289 images are linked to at least one question-answer pair, while 715 images are unused. After removing duplicate image-question-answer triplets across splits, the final dataset comprises 32,632 unique question-answer pairs over 4,289 images. We adopt the official split, consisting of 19,654 training, 6,259 validation, and 6,719 testing question-answer pairs.

**CXR-LT2024**[40] is a large-scale long-tailed multi-label classification benchmark for chest X-rays, released as part of the Long-Tailed Medical Classification Challenge 2024. We adopt the official data splits for Task 2. The training and validation sets include 258,871 and 39,293 images spanning 40 label categories. We compare model performance against expert clinicians using the 406 testing images from Task 2 that provide high-quality, expert-verified annotations across 26 label categories. The demographic attributes of patients in CXR-LT2024 were extracted from the MIMIC-IV dataset[54]. Patients that could not be linked to the MIMIC-IV data were not considered when assessing demographic bias.

**ISIC2018**[32,33] is a dermoscopic image dataset introduced as part of the melanoma detection challenge. We utilize Task 3 (disease classification) to evaluate long-tailed classification performance. Each image is annotated with one of seven disease categories. Following the official split, the training, validation, and testing sets contain 10,015, 193, and 1,512 images, respectively.

**Data Availability**

All datasets used in this study are publicly available and can be obtained from respective open-source repositories. Specifically, the links to the datasets are provided as follows:

PubMedVision: https://huggingface.co/datasets/FreedomIntelligence/PubMedVision,

IMed-361M: https://huggingface.co/datasets/General-Medical-AI/IMed-361M,

MIMIC-CXR v2: https://physionet.org/content/mimic-cxr-jpg/2.0.0/,

PMC-VQA: https://huggingface.co/datasets/RadGenome/PMC-VQA,

VQA-RAD: https://huggingface.co/datasets/flaviagiammarino/vqa-rad,

PathVQA: https://huggingface.co/datasets/flaviagiammarino/path-vqa,

CXR-LT2024: https://physionet.org/content/cxr-lt-iccv-workshop-cvamd/1.1.0/,

ISIC2018: https://challenge.isic-archive.com/data/#2018.

**Code Availability**

Our codes will be made available at https://github.com/lich0031/BiasCareVL.

**Acknowledgements**

This research was partly supported by the National Natural Science Foundation of China (No. U22A2040, No. 62502511), Shenzhen Medical Research Fund (No. B2402047), Natural Science Foundation of Guangdong Province (No. 2026A1515011821), National Key R&D Program of China (No. 2023YFA1011400), Key Laboratory for Magnetic Resonance and Multimodality Imaging of Guangdong Province (No. 2023B1212060052), and Youth Innovation Promotion Association of Chinese Academy of Sciences.